\documentclass[11pt]{article}
\usepackage{xcolor}
\usepackage{booktabs}
\usepackage{multirow}
\usepackage{colortbl}
\usepackage{xcolor}
\usepackage{graphicx}  % in preamble
\usepackage{subcaption}
\definecolor{lightgray}{gray}{0.9}
\usepackage{amsmath}
\usepackage{soul}
\usepackage{arydshln}
\usepackage{enumitem}
\NewDocumentCommand{\hung}
{ mO{} }{\textcolor{cyan}{\textsuperscript{\textit{Hung}}\textsf{\textbf{\small[#1]}}}}

\usepackage[dvipsnames]{xcolor}

\NewDocumentCommand{\oanh}
{ mO{} }{\textcolor{yellow}{\textsuperscript{\textbf{oanh}}\textsf{{\small[#1]}}}}
\NewDocumentCommand{\khoa}
{ mO{} }{\textcolor{red}{\textsuperscript{\textbf{Khoa}}\textsf{{\small[#1]}}}}
\definecolor{kh}{HTML}{168aff}

\usepackage[preprint]{acl}

\usepackage{times}
\usepackage{latexsym}

\usepackage[T1]{fontenc}
\usepackage[utf8]{inputenc}

\usepackage{microtype}

\usepackage{inconsolata}

\usepackage{graphicx}

\title{Look Before You Zoom: Adaptive Routing for the Resolution-Context Trade-off in Visual RAG}

\author{
Oanh N. Tran\textsuperscript{1} \quad
Thanh Quoc Hung Le\textsuperscript{1} \quad
Oscar Chew\textsuperscript{2} \quad
Kuan-Hao Huang\textsuperscript{*,2}
\\
\textbf{Khoa D. Doan}\textbf{\textsuperscript{*,1}}
\\
\textsuperscript{1}VinUni-Illinois Smart Health Center, VinUniversity \quad
\textsuperscript{2}Texas A\&M University
\\
\{26oanh.tn, 21hung.ltq, khoa.dd\}@vinuni.edu.vn
\\
\{oscarchew, khhuang\}@tamu.edu
\\
\textsuperscript{*}Corresponding authors
}
\begin{document}
\maketitle
\begin{abstract}
Vision-Language Models (VLMs) struggles as query-relevant objects become smaller. To address this, recent training-free approaches dynamically retrieve and zoom into local image regions. However, we show that indiscriminately applying retrieval ignores a critical vulnerability: the \textit{resolution--context trade-off}. Patch-based zooming recovers details for small targets, but can split large objects and destroy global spatial context; attention-based retrieval better preserves large objects, but remains less reliable on tiny details; and global perception is often fastest when retrieval is unnecessary. Motivated by these failure modes, we introduce ViRGo (Visual Retrieval or Global Perception), a lightweight framework that formulates visual retrieval as an adaptive routing problem. ViRGo estimates object scale from the VLM’s intrinsic localization heads during the initial forward pass and combines it with semantic token confidence to select between global perception, patch-based retrieval, and attention-based retrieval with minimal additional computation. Experiments across multiple VQA benchmarks and object-size groups show that ViRGo improves the accuracy–efficiency trade-off: it matches patch retrieval on small details, leverages attention-based retrieval for larger objects, and reduces inference time by routing to the global baseline when zooming is unnecessary.

\end{abstract}

\section{Introduction}
% \todo{Add MORE REFERENCES}

Vision-Language Models (VLMs) have recently achieved strong performance in visual question answering (VQA) by jointly reasoning over images and text \citep{liu2023llava, liu2023improvedllava,liu2024llavanext, qwen3technicalreport}. However, their ability to handle objects at varying scales remains limited. While VLMs reliably capture large, salient objects, fixed-resolution visual encoders often dilute fine-grained features, causing small but relevant details to be overlooked \citep{zhang2024exploring}.

A growing line of work addresses this issue through training-free visual retrieval, which dynamically zooms into local image regions to enhance resolution \citep{zhou2025look, wang2025retrieval}. These approaches are particularly effective for small objects, but we observe that applying retrieval indiscriminately introduces a key limitation, which we term the \textit{resolution--context trade-off} (Sec. \ref{sec:resolution_context}). While local zooming improves fine-detail perception, it can disrupt global spatial context, especially for large objects. This fragmentation weakens the model’s ability to reason about object relationships and can degrade performance. Moreover, always-on retrieval incurs significant computational overhead (details in Table.~\ref{tab:Total_run_time}). Together, these suggest that existing visual retrieval methods leave a more basic question unanswered: \textit{Should retrieval be applied at all?}
% These suggesting that neither purely global nor uniformly retrieval-based strategies are consistently optimal.

To better understand this behavior, we analyze three common inference paradigms: global perception, patch-based retrieval \citep{wang2025retrieval}, and attention-based retrieval \citep{zhou2025look}. 
Our analysis reveals that these approaches behave differently across object scales Fig.~\ref{fig:accuracy_comparison}. 
Patch-based retrieval is most effective for small targets, but can fall below global perception when large objects are fragmented. In contrast, attention-based retrieval can preserve spatial context while suppressing irrelevant background noise, allowing it to outperform the global baseline in some large-object settings. These observations suggest that different strategies are optimal under different conditions. Motivated by this, we frame visual reasoning as a routing problem: \textbf{\textit{When should a model rely on global perception, and when should it perform visual retrieval?}} This decision is non-trivial as object scale and task difficulty are unknown at inference time.

To address this, we propose \textbf{\textit{ViRGo}} (\textbf{Vi}sual \textbf{R}etrieval or \textbf{G}l\textbf{o}bal Perception Router), a lightweight routing module that adaptively selects between global inference, attention-based cropping, and patch-based retrieval. ViRGo operates by extracting a zero-shot geometric proxy for object scale directly from the VLM’s intrinsic signals (details in Section~\ref{sec:feature_extraction}).

More broadly, our work extends retrieval-augmented generation (RAG) to the visual domain by introducing \textit{visual RAG routing}: deciding not only what to retrieve, but whether and how to retrieve. Our contributions are:

\begin{itemize}
    \item \textbf{Identification of the Resolution--Context Trade-off:} In Sec.~\ref{sec:resolution_context}, we analyze how visual retrieval strategies behave across object scales. Patch-based retrieval recovers fine-grained details but may fragment large objects, while global perception and attention-based retrieval better preserve context yet are less reliable on small targets.
    
    \item \textbf{Dynamic Scale Estimation via Intrinsic Signals:} We propose a mechanism for estimating perceptual difficulty by combining implicit bounding boxes from the VLM’s intrinsic localization heads with model confidence to construct a geometric proxy for object scale.
    
    \item \textbf{Adaptive Accuracy--Efficiency Routing:} We introduce ViRGo, a lightweight visual RAG router that achieves a better accuracy--latency trade-off across multiple VQA datasets by selectively applying retrieval only when necessary, consistently operating above existing methods on the Pareto frontier (Fig.~\ref{fig:router_pareto_combined}).
\end{itemize}

\section{Related Work}

\subsection{Perception Challenges in VLMs}

VLMs typically rely on fixed-resolution visual encoders (e.g., $336 \times 336$ in LLaVA), which can lead to information loss when processing high-resolution images. This limitation is particularly critical for visual question answering (VQA), where small but semantically important objects may be overlooked due to the perception limitation of VLMs ~\cite{zhang2025mllms}.
% due to feature dilution~\cite{zhang2025mllms} .

To address this issue, prior work has explored several directions. \textbf{High-resolution visual encoders}~\citep{wei2024vary, lu2024deepseek, Li2024MiniGeminiMT, Ge2024ConvLLaVAHB} enhance perception by incorporating stronger visual backbones (e.g., SAM~\cite{Kirillov2023SegmentA}, ConvNeXt~\cite{Liu2022ACF}) or architectural modifications such as cross-attention and adapters. While these approaches improve fine-grained understanding, they often require substantial architectural changes and additional training. \textbf{Input augmentation methods}~\cite{liu2024llavanext, li2024llavaonevisioneasyvisualtask, Chen2024DragonflyMZ, Zhang2024BeyondLD} divide the input image into multiple patches, encode each patch independently, and aggregate the representations for downstream reasoning. Although this strategy improves local detail and it highly effective, this scaling naturally increases computational cost and presents challenges in maintaining global spatial coherence. 

\subsection{Retrieval-Augmented Generation in Multimodal Systems}
Retrieval-augmented generation (RAG) has been widely studied in both textual and multimodal settings~\cite{Chang2021WebQAMA, Han2017AutomaticSF, Xia2024MMedRAGVM, Yu2024VisRAGVR}, where external information is retrieved to enhance model predictions. Recent work extends this paradigm to vision-language tasks by retrieving query-relevant image regions instead of textual documents~\cite{Wang2024DivideCA, shen2024zoomeye, vstar}. 
Methods such as~\cite{wang2025retrieval, zhou2025look} dynamically zoom into informative regions using structured search strategies that better preserve object relationships, leading to improved performance on small-object and relational queries. However, these approaches inherently introduce a trade-off: while zooming enhances local detail, it may lose the global context while incurring additional latency due to iterative retrieval. In fact, in the LLM domain, universal application of retrieval is known to even degrade accuracy due to noisy retrieval~\cite{guo2025routerag, jeong2024adaptive}, suggesting opportunities for adaptive decision strategies.  

\paragraph{Our Perspective.} We formulate visual perception as a routing problem. Rather than committing to a single inference strategy, we ask: \textbf{\textit{When should a model rely on global perception, and when should it perform visual retrieval?}} Specifically, we adaptively choose between global perception, attention-based cropping, and patch-based retrieval as the decision making process on an input. This extends RAG in the visual domain by introducing a key decision dimension: not only \textbf{\textit{how}} to retrieve, but \textbf{\textit{whether and how}} to retrieve under the resolution--context trade-off.

% \paragraph{Our Perspective.}
% We instead view this as a routing problem. Rather than committing to a single inference strategy, we ask: \textit{when should a model rely on global perception, and when should it perform visual retrieval?}

% We propose to address this by adaptively selecting among global perception, attention-based cropping, and patch-based retrieval based on the input. This perspective extends retrieval-augmented generation (RAG) to the visual domain by introducing a key decision dimension: not only \textit{how} to retrieve, but \textit{whether and how} to retrieve under the resolution--context trade-off.

\section{Motivations: Understanding the Resolution--Context Trade-off}
\subsection{Background}

\begin{figure*}[t]
    \centering
    \includegraphics[width=0.9\linewidth]{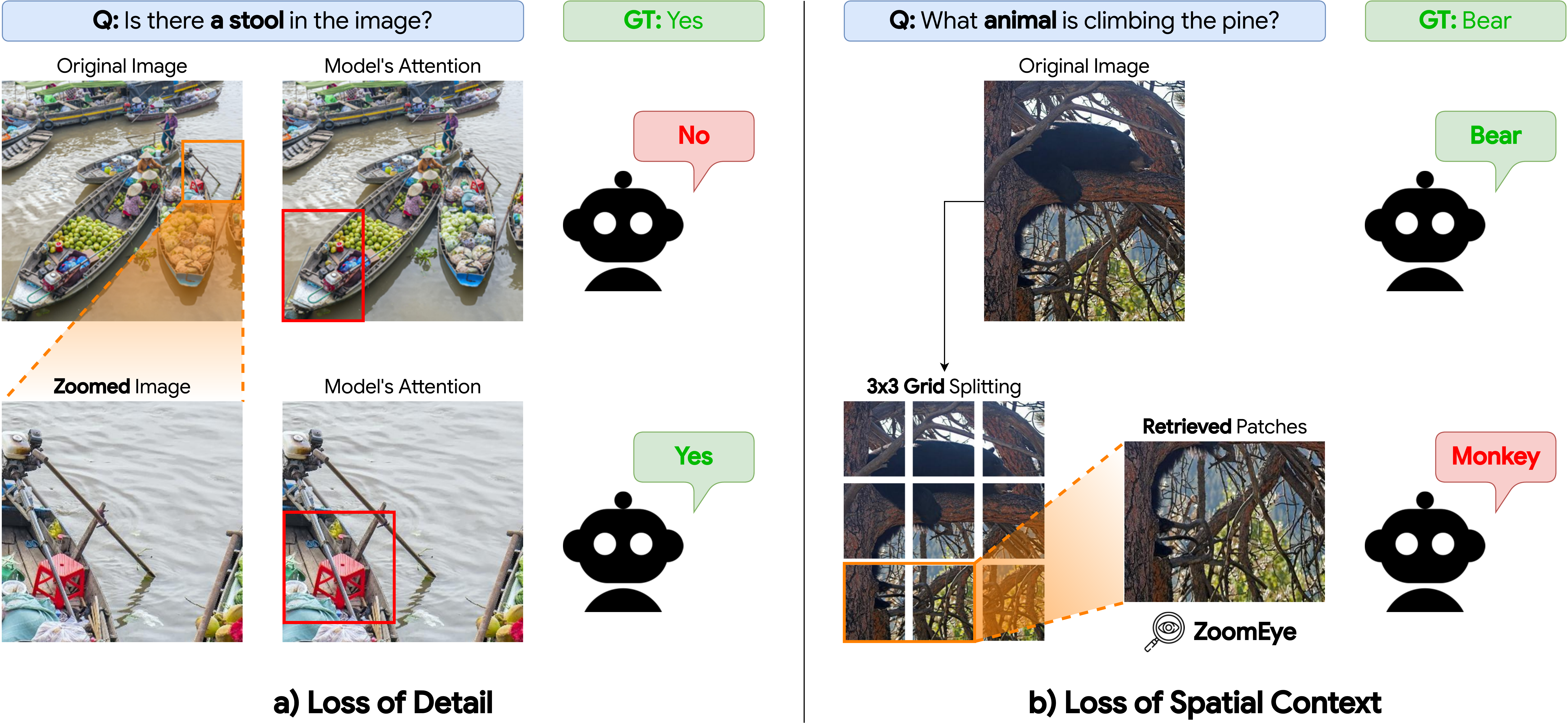}
\caption{
Qualitative examples of the resolution--context trade-off. 
(a) \textbf{Loss of detail}: For small objects, global perception fails because the target occupies only a few visual tokens, while zooming into the relevant region restores fine-grained detail and yields the correct answer. 
(b) \textbf{Loss of spatial context}: For large objects, patch-based retrieval can split the target across regions, losing global structure and surrounding context, leading to incorrect prediction.}
    \label{fig:combined_failures}
\end{figure*}

A typical VLM consists of three main components: (1) a visual encoder (e.g., CLIP~\citep{radford2021learning}), (2) a projector, and (3) an LLM. The visual encoder processes an image by dividing it into a grid of fixed-size patches and extracting features for each patch. The projector then maps these visual features into the input space of the LLM, allowing the model to reason over both image and text. This design limits how the model handles objects at different sizes. Because the image is resized to a fixed resolution (e.g., $336 \times 336$), small objects occupy very few patches and can be overlooked by the model (see left panel of Fig.~\ref{fig:combined_failures}). Retrieval-based methods address this issue by selecting image regions using an external retriever and zooming in before passing them to the model \cite{wang2025retrieval}. Nevertheless, zoomed-in patches may cover only part of the object and lose surrounding context, making it harder for the model to understand the full scene. In this section, we investigate this resolution-context trade-off when universally applying either global perception or retrieval-based zooming.
\subsection{Resolution--Context Trade-off}
\label{sec:resolution_context}
% \hl{Just Update (full section)}
To better understand the resolution--context trade-off, we analyze three representative inference paradigms in VLM-based VQA:
\begin{enumerate}
    \item \textbf{Global perception} (baseline), which processes the entire image at a fixed resolution;
    \item \textbf{Attention-based retrieval}, which adaptively retrieves and zooms into salient regions using model's attention (we use ViCrop \citep{zhang2025mllms} as a representative).
    \item \textbf{Patch-based retrieval}, where the model retrieves and zooms into local regions (we use RAP \citep{wang2025retrieval} as a representative).

\end{enumerate}

\begin{figure}[t]
    \centering
        \centering
\includegraphics[width=0.48\textwidth]{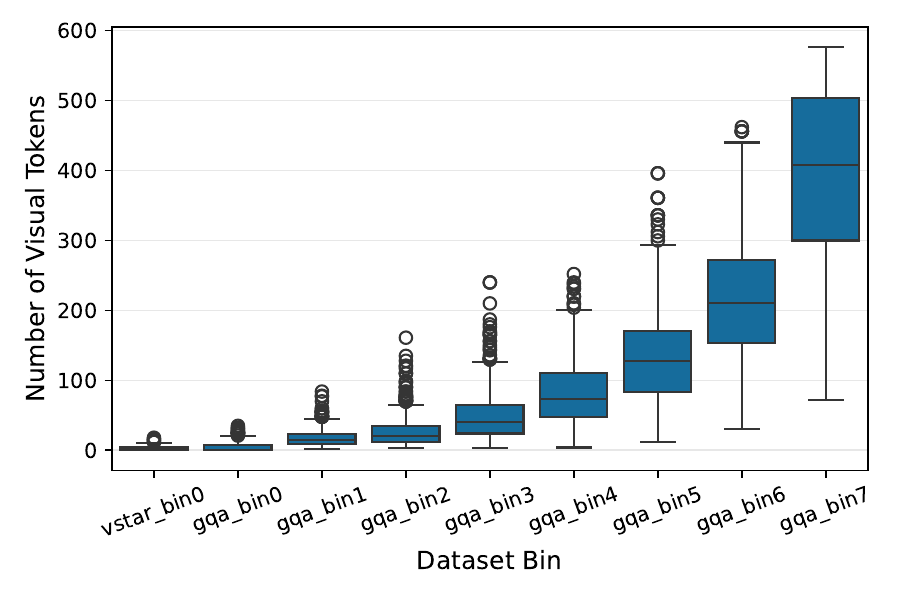}
        \caption{Object size distribution}
        \label{fig:token_size_dist}
\end{figure}

\noindent \textbf{Setup.} We conduct our analysis on two complementary benchmarks. \textbf{GQA} \citep{hudson2019gqa} evaluates compositional visual reasoning over real-world scenes with diverse object scales, while \textbf{V$^*$ Bench} \cite{vstar} provides high-resolution images with fine-grained, detail-sensitive queries. Both datasets include ground-truth bounding boxes, enabling us to estimate object scale for controlled analysis. 
% To study behavior across different object sizes, we partition GQA into bins based on the relative size of the target object. \hung{We need to clearly define this procedure, maybe Appendix (then point to Appendix).}This allows us to systematically evaluate how each inference strategy performs as object scale varies. Figure~\ref{fig:combined} shows the distribution of object sizes and image resolutions across these bins.
We partition GQA into bins based on the relative size of the target object. Figure~\ref{fig:token_size_dist} shows the distribution of object sizes and image resolutions across these bins.

\begin{figure}[!ht]
    \centering
    \begin{subfigure}[h]{0.45\textwidth}
        \centering
        \includegraphics[width=\textwidth]{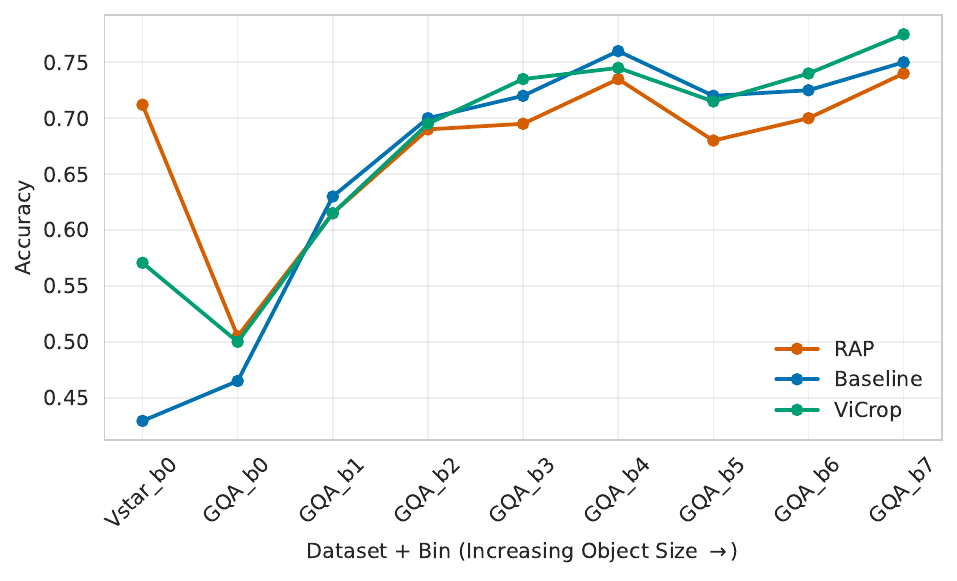}
        \caption{Accuracy comparison}
        \label{fig:accuracy_comparison}
    \end{subfigure}
    \hfill
    \begin{subfigure}[h]{0.45\textwidth}
        \centering
        \includegraphics[width=\textwidth]{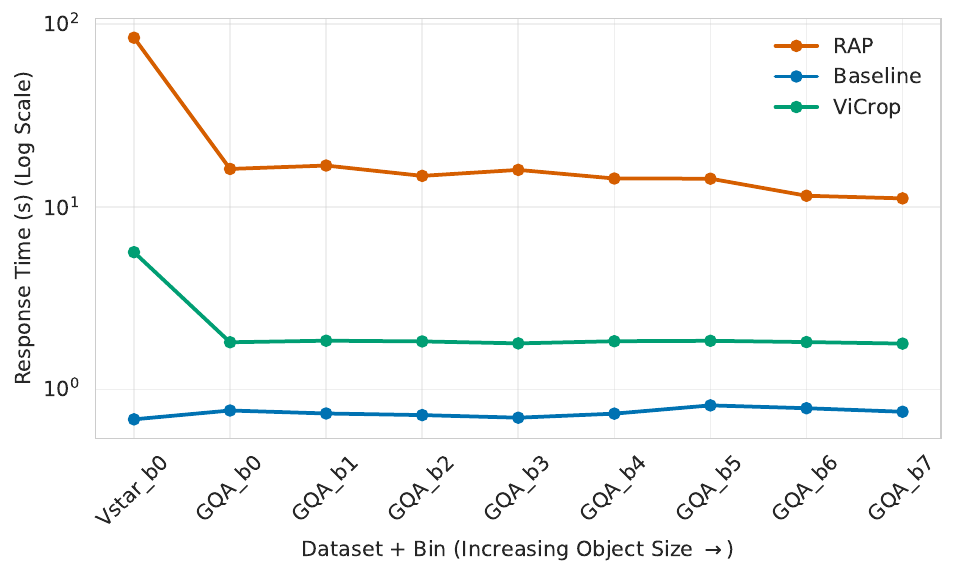}
        \caption{Response time (log scale)}
        \label{fig:response_time_comparison}
    \end{subfigure}
    \caption{
    Comparison of RAP (orange), ViCrop (green) and Baseline (blue) across datasets. 
    Left: accuracy across bins. Right: response time per sample (log scale). 
    Higher bins correspond to larger ground-truth regions.
    }
    \label{fig:combined_graphs}
\end{figure}

\subsection{Observations: Scale-Dependent Failure Modes}

We reveal two complementary failures that explain the trade-off between image resolution and spatial context. The first failure occurs for small objects, which occupy only a few visual tokens and are therefore difficult for standard global inference and attention-based cropping to detect. The second failure occurs for large objects, which can be divided across multiple zoomed-in patches, causing patch-based retrieval to lose surrounding spatial context and object-level relationships. We study these failures both qualitatively and quantitatively by comparing global perception, attention-based cropping, and patch-based retrieval across object scales.
\paragraph{Loss of Detail in Small Objects.} 
The left panel of Figure~\ref{fig:combined_failures} shows that global perception often fails on small, localized targets. Since VLMs process images at a fixed resolution, small objects occupy only a tiny fraction of the visual tokens and their features become diluted by surrounding context. Consequently, the model fails to localize the target and produces hallucinated answers. In contrast, retrieval-based zooming enlarges the target region, allowing the visual encoder to recover fine-grained details and correctly answer the query.

\paragraph{Loss of Context in Large Objects.} 
The right panel of Figure~\ref{fig:combined_failures} shows the opposite failure mode for large objects. Fixed-grid retrieval fragments large objects across multiple patches, removing the global spatial context required for recognition. As a result, the model only observes partial object regions and generates locally plausible but incorrect predictions. This demonstrates that excessive patch-based retrieval can harm recognition when global context is essential.

\paragraph{Quantitative Trends Across Object Scales.}
We further provide a large-scale analysis of the resolution--context trade-off in Figure~\ref{fig:combined_graphs}. Figure~\ref{fig:accuracy_comparison} compares accuracy across V*Bench and GQA bins, where higher GQA bins correspond to larger target regions. RAP performs best on V*Bench, where queries often require fine-grained perception of very small targets. However, as object size increases, patch-based retrieval becomes less consistently beneficial and often falls below the global baseline on GQA bins. This trend supports the qualitative observation above: zooming helps when the target is too small to perceive globally, but it becomes harmful when the target is already visible and retrieval fragments the object or removes useful context.

Figure~\ref{fig:response_time_comparison} shows that this accuracy trade-off is also tied to efficiency. RAP consistently incurs the largest response time because it requires iterative retrieval and multiple zoomed-in evaluations. In contrast, the global baseline is the fastest, and attention-based cropping provides an intermediate option that can suppress irrelevant background while avoiding some of the fragmentation caused by rigid patch retrieval. Starting from the larger GQA bins, processing the full image is often both more accurate and substantially faster than always applying patch-based retrieval.

\begin{figure*}[!ht]
    \centering
    \includegraphics[width=0.95\linewidth]{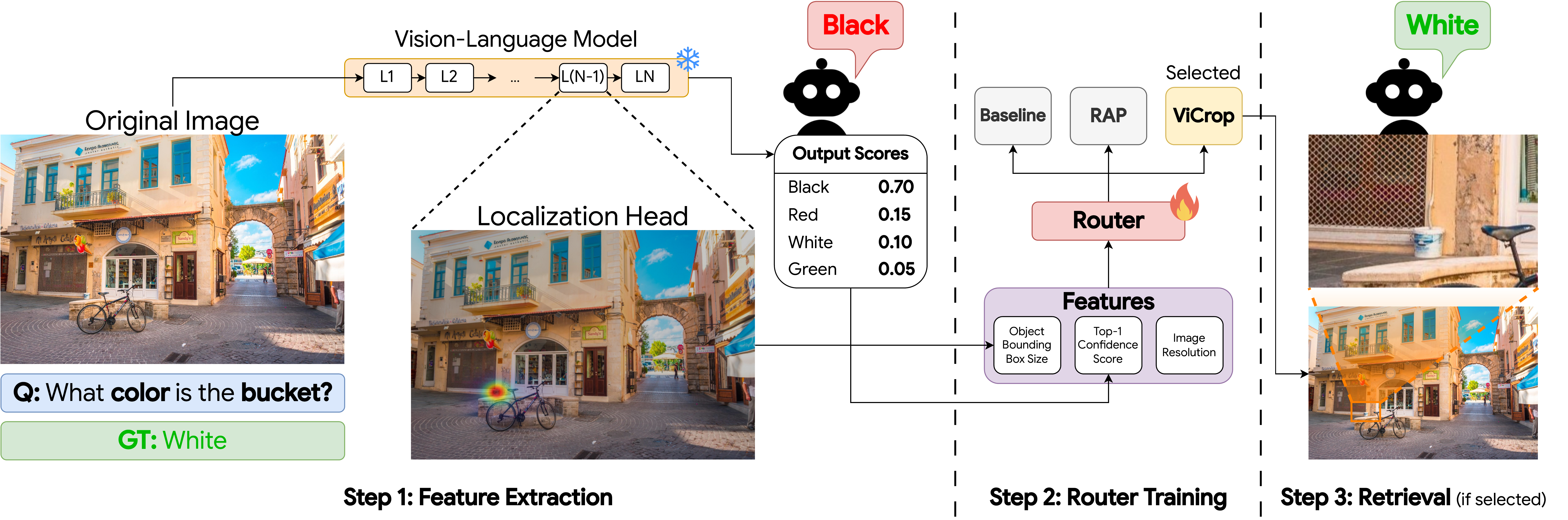}
    \caption{
    \textbf{Overview of ViRGo.}
    Given an input image and question, ViRGo first performs a global forward pass through the $N$-layer vision-language model to extract zero-shot routing signals. These include an implicit object bounding-box size from localization heads, the top-1 confidence score, and the image resolution. The extracted features are passed to a lightweight router, which selects the most suitable perception strategy among global inference, patch-based retrieval with RAP, and attention-based cropping with ViCrop. In this example, the global model initially predicts ``Black,'' while the router selects ViCrop to focus on the relevant image region and recover the correct answer, ``White.''
    }
    \label{fig:main_figure}
\end{figure*}

% The analysis suggests that optimal visual processing should be selected dynamically at inference time rather than fixed a priori.
The resolution--context trade-off shows that no single perception strategy is uniformly optimal, motivating an adaptive mechanism that selects the strategy best suited to each input.
This perspective is also consistent with human visual processing, where the visual system adaptively shifts between global scene understanding and focused attention depending on the task and visual complexity~\cite{torralba2006contextual}. In this section, we describe ViRGo, an adaptive router that decides whether the VLM should rely on Visual Retrieval or Global Perception during inference time. The overall workflow of ViRGo is illustrated in Figure~\ref{fig:main_figure}: an initial global forward pass provides zero-shot routing features, which are then used by a lightweight router to select the most suitable perception strategy among global perception, RAP, and ViCrop. 

% To develop ViRGo, we have two big questions to answer: 
% \begin{enumerate}
%     \item How to know the size of object without ground-truth? 
%     \item How to automatically define the threshold? 
% \end{enumerate}

\subsection{Zero-Shot Feature Extraction}\label{sec:feature_extraction}
% \todo{Add figures to show diffuse attention case and over-confidence of model confidence}
% , often referred to as \textit{localization heads}

To dynamically decide the visual processing strategy, the VLM must be able to estimate the physical scale of the queried object without relying on ground-truth annotations. Inspired by recent interpretability studies, where spatial grounding is implicitly encoded within a sparse subset of the LLM's attention heads~\cite{kang2025your, zhang2025mllms}, we propose to aggregate the attention weights from these heads, often called localization heads, over the visual tokens during the initial forward pass, and extract a soft localization map. 
% In this work, we convert this attention map into an implicit bounding box ($bb\_size$) \hung{this sentence is unclear. what is "this attention map"?}. 
% Crucially, because these localization heads activate during the baseline global inference step, extracting this spatial extent incurs effectively zero computational overhead.
% These specific heads exhibit strong, consistent attention weights that spatially align with the semantic objects currently being generated in the text, even though the VLM is trained purely on text-generation objectives. 
The extracted features are as follows:
% To determine the optimal strategy without relying on ground-truth annotations or expensive external object detectors, we extract two intrinsic zero-shot signals from the MLLM's initial global forward pass:
\begin{enumerate}
    \item \textbf{Implicit Object Scale ($bb\_size$)}: The spatial bounding box size extracted directly from the VLM's localization heads \citep{kang2025your}, serving as a geometric proxy for the target object's physical scale within the image. 
    These specific heads exhibit strong, consistent attention patterns that spatially align with the semantic objects currently being generated in the text, even though the VLM is trained purely on text-generation objective. 
    % By aggregating the attention weights from these localization heads over the visual tokens during the initial forward pass, we can extract a soft localization map. In this work, we convert this attention map into an implicit bounding box ($bb\_size$) \hung{this sentence is unclear. what is "this attention map"?}. 
    % Crucially, because these localization heads activate during the baseline global inference step, extracting this spatial extent incurs effectively zero computational overhead.
    
    \item \textbf{Semantic Confidence ($top\_token\_prob$)}: 
    The probability of the highest-scoring initially generated token, used as a proxy for the model's internal certainty. 
    This signal complements object scale: even when the estimated object size is large, low confidence may indicate that the model is uncertain or attending to the wrong visual evidence. 
    In such cases, retrieval may still be beneficial.

    \item \textbf{Source Image Size ($image\_resolution$)}: The raw resolution of the input image. 
    While small objects often benefit from zoom-in retrieval, this assumption can fail when the source image itself has insufficient resolution. 
    In such cases, aggressive cropping may produce low-quality patches that fall below the visual encoder's effective recognition threshold, making retrieval detrimental. 
    Image resolution therefore helps the router identify when zoom-in retrieval should be avoided.
\end{enumerate}

\section{Proposed Method: ViRGo}

We formulate adaptive perception routing using three independent binary decision models corresponding to: \textbf{0:} Global Baseline, \textbf{1:} Attention-based Retrieval (ViCrop), and \textbf{2:} Patch-based Retrieval (RAP).
Each lightweight XGBoost router predicts the suitability of its corresponding strategy from the extracted routing features, and its output is further calibrated into a comparable confidence score. During inference, the final perception strategy is selected based on the highest calibrated prediction score under a minimal-computation preference, enabling routing decisions with minimum additional overhead. Detailed implementation is in App.~\ref{detailed_implementation}.

\section{Experiments}
\subsection{Experimental Setup}

\paragraph{Datasets.}
We evaluate the methods on multiple VQA benchmark, including subset of GQA \cite{hudson2019gqa}, V$^*$ Bench \cite{vstar}, HR-Bench 4K \cite{Wang2024DivideCA}, subset of TextVQA \cite{singh2019towards}. For datasets where no official split is provided, we create a train/dev/test split and report results on the held-out test set. The number of test samples for each dataset is summarized in Table~\ref{tab:test_samples}. These datasets cover a wide range of object scales.
Details on training, validation, and test splits are provided in the Appendix.
\begin{table}[t]
\centering
\small
\setlength{\tabcolsep}{4pt}
\renewcommand{\arraystretch}{1.05}

\begin{tabular}{lcccc}
\toprule
\textbf{Dataset} & \textbf{N (all)} & \textbf{N (train)} & \textbf{N (dev)} & \textbf{N (test)} \\
\midrule
GQA bin-0 & 200 & 40 & 32 & 128 \\
GQA bin-1 & 200 & 40 & 32 & 128 \\
GQA bin-2 & 200 & 40 & 32 & 128 \\
GQA bin-3 & 200 & 40 & 32 & 128 \\
GQA bin-4 & 200 & 40 & 32 & 128 \\
GQA bin-5 & 200 & 40 & 32 & 128 \\
GQA bin-6 & 200 & 40 & 32 & 128 \\
GQA bin-7 & 200 & 40 & 32 & 128 \\
HR-Bench4K & 800 & 160 & 128 & 512 \\
TextVQA & 200 & 40 & 32 & 128 \\
V*Bench & 191 & 38 & 30 & 123 \\

\midrule
\textbf{Total} & \textbf{2791} & \textbf{558} & \textbf{446} & \textbf{1787} \\
\bottomrule

\end{tabular}

\caption{Detailed Dataset statistics for training and evaluation splits.}
\label{tab:test_samples}
\end{table}

\paragraph{Label Construction.} For each sample, we evaluate all three strategies and assign the label representing the method with both the correct answer and the least computation, ensuring a preference efficiency. If all methods fail, the label defaults to global visual processing.

% \begin{itemize}
%     \item If global perception is correct, assign label 0;
%     \item Else if attention-based cropping is correct, assign label 1;
%     \item Else if patch-based retrieval is correct, assign label 2;
%     \item If all methods fail, assign label 0.
% \end{itemize}
% This ensures the preference for a more efficient method when possible.

% this is not necessary \paragraph{Overview.}
% Given three signals extracted from the model—object size ($s$), token confidence ($p$), and image resolution ($r$)—we train a lightweight router to select the most suitable perception strategy. We formulate this as a 3-class classification problem:
% \begin{itemize}
%     \item 0: Global perception
%     \item 1: Attention-based cropping (ViCrop)
%     \item 2: Patch-based retrieval (RAP)
% \end{itemize}

% \paragraph{Oracle Label Construction.}
% Since ground-truth routing labels are not available, we construct labels using a simple cost-aware rule. For each sample, we evaluate all three strategies and assign the label based on the cheapest method that produces the correct answer:
% \begin{itemize}
%     \item If global perception is correct, assign label 0;
%     \item Else if attention-based cropping is correct, assign label 1;
%     \item Else if patch-based retrieval is correct, assign label 2;
%     \item If all methods fail, assign label 0.
% \end{itemize}
% This ensures the model prefers faster methods when possible and avoids unnecessary computation.

\paragraph{VLM Backbones.}
We use LLaVA-v1.5-7B and LLaVA-v1.5-13B (fixed resolution) and LLaVA-OneVision-0.5B (LLaVA-ov-0.5B) and Qwen3-VL (dynamic resolution) in our experiments. 

\paragraph{Baselines.}
We compare our approach with RAP (patch-based retrieval) and ViCrop (attention-based retrieval), besides the original VLM visual processing, i.e., global perception (baseline). All methods follow the original paper’s experimental setup, including the instruction to generate a single token for evaluation consistency. For routing, we also include a random router as a baseline.

\paragraph{Evaluation Metrics.}
We report Accuracy (Exact Match, \%) and total inference time (seconds). For our method, the reported time includes both routing and response generation.

% \subsection{Main Experiment}

% \paragraph{Retrieval Methods Implementation Details.}
% All methods follow the original paper’s setup, including the instruction to generate a single token for evaluation consistency. We ensure fair comparison by matching the base generation setting (without zoom) across all methods.

% \subsection{Main Results}
\subsection{Performance Results}

\begin{table*}[!ht]
\centering
\small
\setlength{\tabcolsep}{3.0pt}
\resizebox{\textwidth}{!}{
\begin{tabular}{lccccccccc}
\toprule
\multirow{2}{*}{\textbf{Method}} 
& \multicolumn{3}{c}{\textbf{Small object (N = 763)}} 
& \multirow{2}{*}{\textbf{Overall}} 
& \multicolumn{2}{c}{\textbf{Medium object (N = 512)}} 
& \multirow{2}{*}{\textbf{Overall}} 
& \multicolumn{1}{c}{\textbf{Large object (N = 512)}} 
& \multirow{2}{*}{\textbf{\textit{Weighted Avg $\uparrow$}}} \\

\cmidrule(lr){2-4}
\cmidrule(lr){6-7}
\cmidrule(lr){9-9}

& \textbf{VsB} 
& \textbf{HR4K} 
& \textbf{G0} 
&  
& \textbf{G1--G3} 
& \textbf{TextVQA} 
&  
& \textbf{G4--G7} 
% & \textbf{Overall} 
& \\

\midrule

\multicolumn{10}{c}{\textit{Fixed Resolution Model}} \\
\midrule

LLaVA-v1.5-7B &
36.6 &
35.0 &
45.3 &
37.0 &
\underline{68.2} &
40.6 &
61.3 &
74.0 &
54.6
\\

\ - ViCrop &
\underline{55.3} &
48.6 &
\underline{48.4} &
49.7 &
\underline{68.2} &
\textbf{50.0} &
\textbf{63.7} &
\underline{74.8} &
\underline{60.9}
\\

\ - RAP &
\textbf{69.1} &
\underline{50.6} &
\underline{48.4} &
\underline{53.2} &
68.0 &
44.5 &
62.1 &
71.1 &
\underline{60.9}
\\

\rowcolor{gray!8}
\ + \textit{Random} &
56.9	&
45.7	&
45.3	&
47.4	&
67.2	&
44.5	&
61.5	&
74.6	&
59.3	
\\

\rowcolor{gray!15}
\ \textbf{+ ViRGo (Ours)} &
\textbf{69.1} &
\textbf{52.5} &
\textbf{53.1} &
\textbf{55.3} &
\textbf{68.8} &
\underline{46.9} &
\underline{63.3} &
\textbf{75.0} &
\textbf{63.2}
\\

\rowcolor{green!12}
\ \textit{* Oracle} &
78.9	&
66.6	&
60.2	&
67.5	&
78.4	&
59.4	&
73.6	&
78.5	&
72.4	
\\

\addlinespace
\hdashline
\addlinespace

LLaVA-v1.5-13B &
42.3 &
41.8 &
45.3 &
42.5 &
\underline{69.0} &
40.6 &
61.9 &
\underline{75.4} &
57.5
\\

\ - ViCrop &
52.9 &
47.3 &
48.4 &
48.4 &
\textbf{69.5} &
\underline{48.4} &
\textbf{64.3} &
\textbf{77.3} &
61.2
\\

\ - RAP &
\textbf{69.1} &
\textbf{58.0} &
50.0 &
\underline{58.5} &
64.8 &
46.9 &
60.4 &
73.2 &
\underline{63.2}
\\

\rowcolor{gray!8}
\ + \textit{Random} &
58.5	&
47.3	&
51.6	&
49.8	&
67.2	&
46.1	&
61.9	&
75.6	&
60.7	
\\

\rowcolor{gray!15}
\ \textbf{+ ViRGo (Ours)} &
\underline{67.5} &
\textbf{58.0} &
\textbf{52.3} &
\textbf{58.6} &
\underline{69.3} &
\underline{46.9} &
\underline{63.7} &
\underline{76.8} &
\textbf{65.3}
\\

\rowcolor{green!12}
\ \textit{* Oracle} &
78.1	&
69.0	&
59.4	&
68.8	&
77.9	&
60.2	&
73.4	&
80.7	&
73.5	
\\
\midrule

\multicolumn{10}{c}{\textit{Dynamic Resolution Model}} \\
\midrule

LLaVA-ov-0.5B &
43.1 &
44.3 &
\textbf{56.3} &
46.1 &
\textbf{65.1} &
\underline{60.9} &
\textbf{64.1} &
\textbf{70.9} &
\underline{58.4}
\\

\ - ViCrop &
46.3 &
43.4 &
49.2 &
44.8 &
60.9 &
\underline{60.9} &
62.3 &
69.7 &
57.0
\\

\ - RAP &
\textbf{48.8} &
\underline{45.7} &
52.3 &
\underline{47.3} &
60.9 &
\textbf{62.5} &
61.3 &
\underline{71.1} &
58.1
\\
\rowcolor{gray!8}
\ + \textit{Random} &
48.8	&
46.5	&
56.3	&
48.5	&
60.9	&
60.2	&
60.7	&
69.3	&
58.0	
\\

\rowcolor{gray!15}
\ \textbf{+ ViRGo (Ours)} &
\underline{48.0} &
\textbf{46.7} &
\textbf{56.3} &
\textbf{48.5} &
\underline{64.8} &
60.2 &
\underline{63.7} &
70.7 &
\textbf{59.2}
\\

\rowcolor{green!12}
\ \textit{* Oracle} &
64.2	&
57.8	&
60.9	&
59.4	&
70.8	&
69.5	&
70.5	&
74.8	&
67.0	
\\

\addlinespace
\hdashline
\addlinespace

Qwen3-VL-2B &
64.2 &
54.1 &
55.5 &
56.0 &
\textbf{66.7} &
\textbf{75.0} &
\textbf{68.8} &
\textbf{72.5} &
64.4
\\

\ -ViCrop &
57.7 &
52.3 &
53.1 &
53.3 &
62.0 &
\underline{72.7} &
64.7 &
67.0 &
60.5
\\

\ -RAP &
\textbf{78.9} &
\underline{70.9} &
53.1 &
\underline{69.2} &
64.6 &
64.8 &
64.6 &
70.1 &
\underline{68.2}
\\
\rowcolor{gray!8}
\ + \textit{Random} &
63.4	&
59.2	&
50.0	&
58.3	&
64.6	&
72.7	&
66.6	&
70.1	&
64.1	
\\

\rowcolor{gray!15}
\ \textbf{+ ViRGo (Ours)} &
\textbf{78.9} &
\textbf{71.1} &
\textbf{56.3} &
\textbf{69.9} &
\underline{66.4} &
\textbf{75.0} &
\underline{68.6} &
\textbf{72.5} &
\textbf{70.2}
\\

\rowcolor{green!12}
\ \textit{* Oracle} &
88.6	&
77.5	&
61.7	&
76.7	&
75.5	&
78.9	&
76.4	&
78.5	&
77.1	
\\

\bottomrule
\end{tabular}
}
\caption{Accuracy (\%) across datasets. G1--G3 and G4--G7 denote averaged GQA bins. The Overall columns denote the weighted average score within each object-size category, while Weighted Avg denotes the weighted average across all datasets in the table. Bold indicates the best performance, while underlining indicates the second-best.}
\label{tab:dataset_wise_grouped}
\end{table*}

\begin{table}[!th]
\centering
\small
\setlength{\tabcolsep}{6pt}
\renewcommand{\arraystretch}{}

\begin{tabular}{lrrrr}
\toprule
\textbf{Method} &
\textbf{\textit{Small}}&
\textbf{\textit{Medium}}&
\textbf{\textit{Large}}&
\textbf{\textit{Total}$\downarrow$} \\
\midrule

\textbf{LLaVA-v1.5-7B} &
541	&
399	&
395	&
1335	
\\

- ViCrop & 
4074	&
1026	&
934	&
6034	
\\

- RAP & 
71506	&
11252	&
6427	&
89185	
 
\\

\rowcolor{gray!12}
+ \textbf{ViRGo} & 
39374	&
3341	&
1911	&
44626	
\\

\hdashline
\textbf{LLaVA-ov-0.5B} &
1080	&
156	&
104	&
1340	
\\

- ViCrop & 
18438	&
1721	&
1080	&
21239	
\\

- RAP & 
67961	&
7913	&
5278	&
81152	
 
\\

\rowcolor{gray!12}
+ \textbf{ViRGo} & 
50182	&
1517	&
682	&
52381	
\\
\bottomrule
\end{tabular}

\caption{Total inference time between methods (s). ViRGo includes both routing and generation time.}
\label{tab:Total_run_time}
\vspace{-15pt}
\end{table}

Across both fixed- and dynamic-resolution backbones, ViRGo consistently achieves the best performance by dynamically choosing between Baseline, ViCrop, and RAP. For LLaVA-v1.5-7B, ViRGo reaches 63.2\%, outperforming the Baseline (+8.6\%) and both ViCrop and RAP ( +2.3\%). On LLaVA-v1.5-13B, ViRGo achieves 65.3\%, improving over the Baseline (+7.8 \%), ViCrop (+4.1 \%), and RAP (+2.1 \%). For Qwen3-VL-2B, ViRGo obtains 70.2\%, surpassing the Baseline (+5.8 \%) and RAP (+2.0 \%), while ViCrop performs below baseline. Even in the challenging case of LLaVA-ov-0.5B, where standalone ViCrop and RAP both underperform the Baseline, ViRGo still improves performance to 59.2\% (+0.8\%) comparing to the Baseline. Overall, these results demonstrate that adaptive routing is consistently more effective than relying on any single fixed inference strategy.

\subsection{Time Complexity Analysis}
Table~\ref{tab:Total_run_time} shows the inference time (in seconds) across all methods. The results for ViRGo already include both the routing overhead and generation time. 
Global perception remains the fastest method across all settings, with ViCrop consistently being the second fastest. In contrast, RAP is the most computationally expensive configuration. ViRGo achieves faster computation compared to RAP while maintaining stronger accuracy; for example, ViRGo's computation is  49.9\% and 35.5\% less than those of RAP on LLaVA-v1.5-7B and LLaVA-ov-0.5B, respectively.

% \begin{itemize}
%     \item \textbf{Baseline and ViCrop:} The Baseline remains the fastest method, while ViCrop is consistently the second fastest.
    
%     \item \textbf{RAP:} RAP is the most computationally expensive configuration, requiring between 50k--94k seconds across models.
    
%     \item \textbf{ViRGo:} ViRGo substantially reduces computation compared to RAP while maintaining higher accuracy. For LLaVA-v1.5-7B, runtime decreases from 89,185s to 44,626s (49.9\% reduction). For LLaVA-ov-0.5B, runtime drops from 81,152s to 52,381s (35.5\% reduction).
% \end{itemize}

On small-object samples,  ViRGo has the highest computation as crop-based reasoning is frequently required. In contrast, ViRGo achieves the largest time savings on medium and large objects by routing many samples directly to the global perception.

\begin{figure*}[!ht]
    \centering
    
    \begin{subfigure}[t]{0.31\textwidth}
        \centering
        \includegraphics[width=\textwidth]{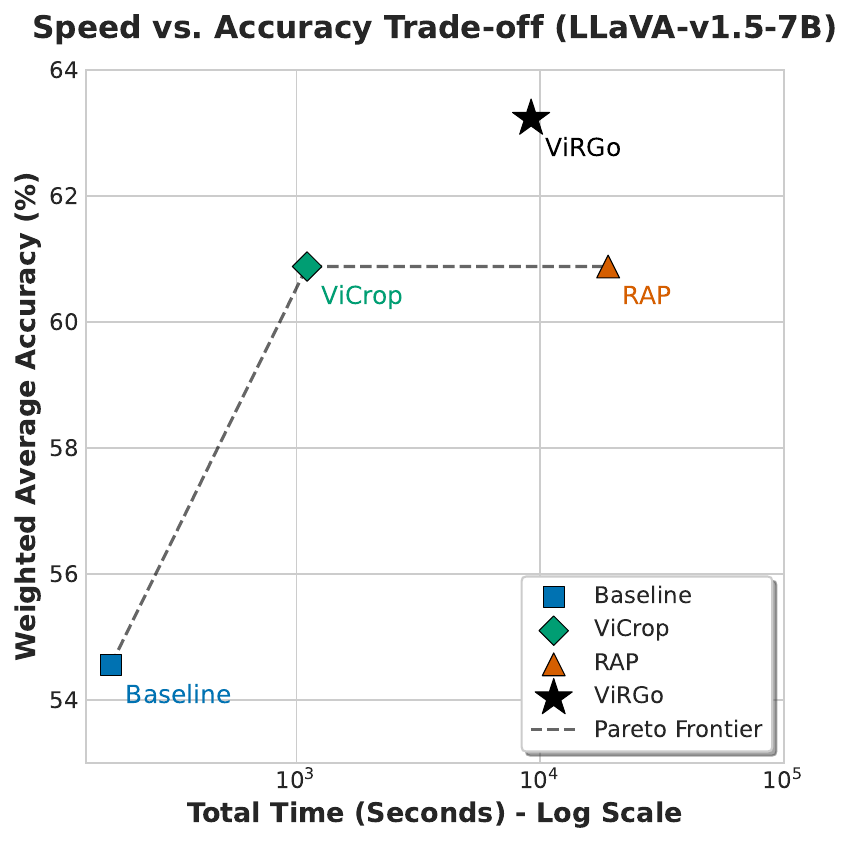}
        \caption{LLaVA-v1.5-7B}
        \label{fig:pareto_7b}
    \end{subfigure}
    \hfill
    \begin{subfigure}[t]{0.31\textwidth}
        \centering
        \includegraphics[width=\textwidth]{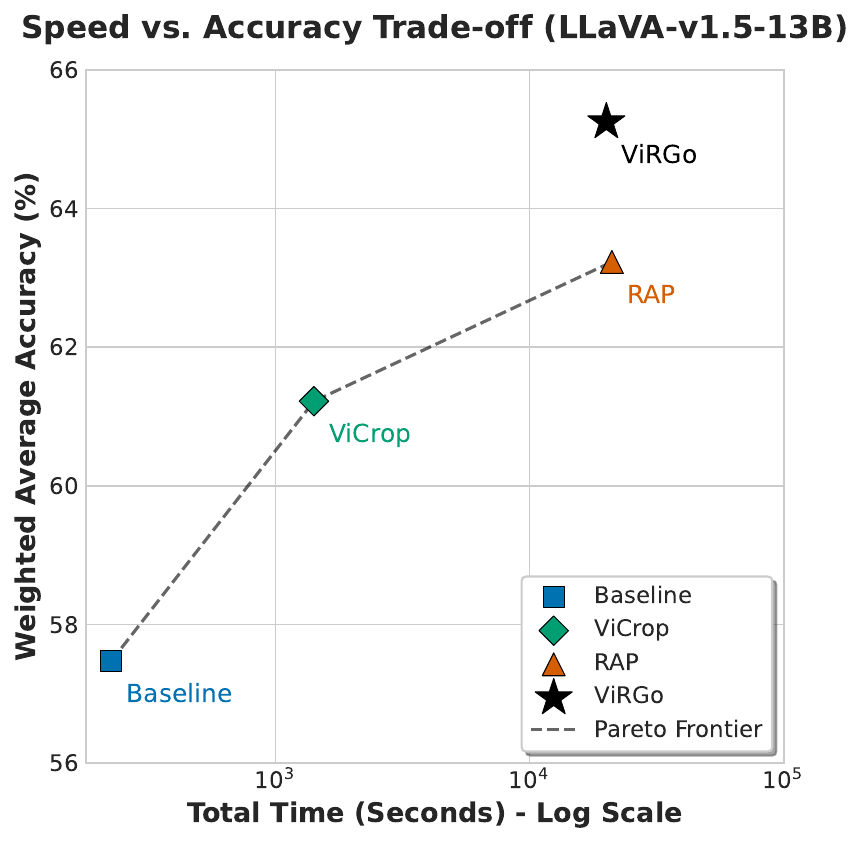}
        \caption{LLaVA-v1.5-13B}
        \label{fig:pareto_13b}
    \end{subfigure}
    \hfill
    \begin{subfigure}[t]{0.31\textwidth}
        \centering
        \includegraphics[width=\textwidth]{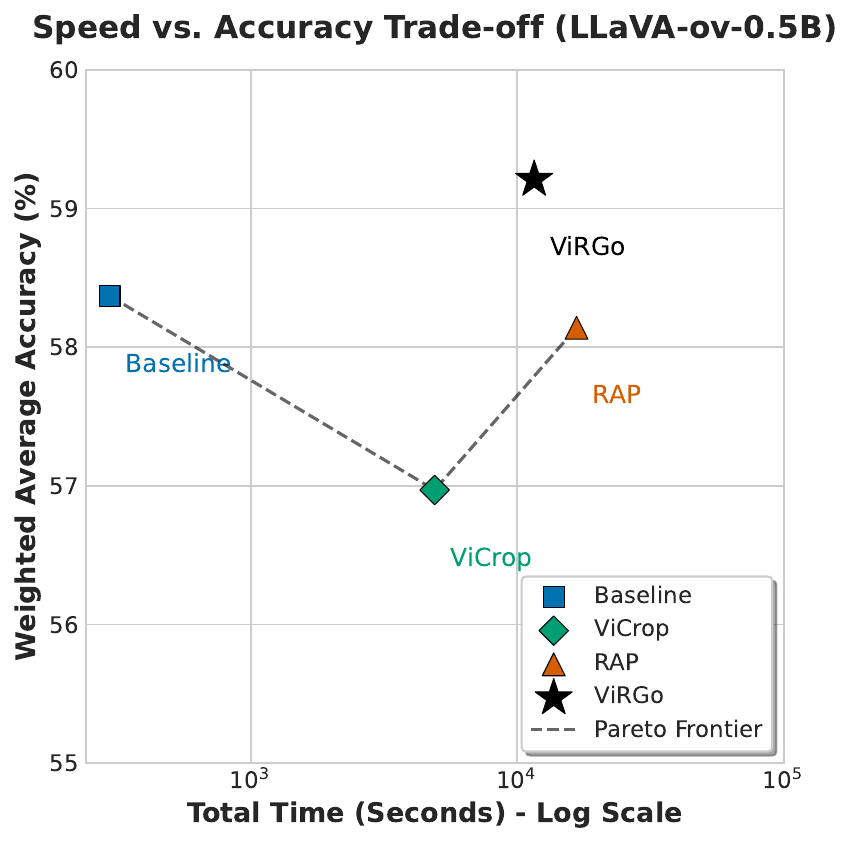}
        \caption{LLaVA-ov-0.5B}
        \label{fig:pareto_ov_05b}
    \end{subfigure}
    
    \caption{
        \textbf{Speed vs. Accuracy Pareto Frontier.} Weighted average accuracy across different datasets compared to total inference time (log scale) for \textbf{(a)} LLaVA-v1.5-7B, \textbf{(b)} LLaVA-v1.5-13B, and \textbf{(c)} LLaVA-ov-0.5B. The dashed line illustrates the Pareto frontier.
    }
    \label{fig:router_pareto_combined}
\end{figure*}

In conclusion, across all evaluated backbones, ViRGo achieves a better accuracy--efficiency trade-off than standalone inference strategies as illustrated in Fig.~\ref{fig:router_pareto_combined}. The router exhibits consistent and expected behavior: relying on RAP for difficult small-object cases while favoring the less expensive global perception or ViCrop for medium and large objects. 
% Consequently, ViRGo achieves a better overall accuracy while significantly reducing the computational cost of exhaustive crop-based inference.

\begin{figure}[t]
    \centering
    \begin{subfigure}[t]{0.48\textwidth}
        \centering
        \includegraphics[width=\textwidth]{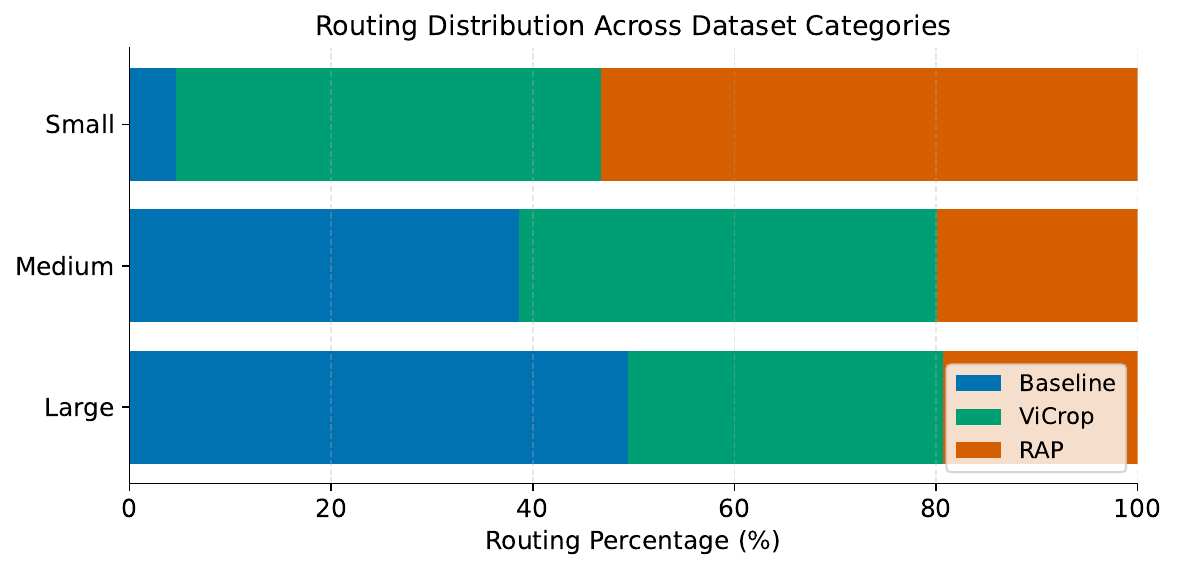}
        \caption{LLaVA-1.5-7B routing distribution}
        \label{fig:llava_routing_distribution}
    \end{subfigure}
    \hfill
    \begin{subfigure}[t]{0.48\textwidth}
        \centering
        \includegraphics[width=\textwidth]{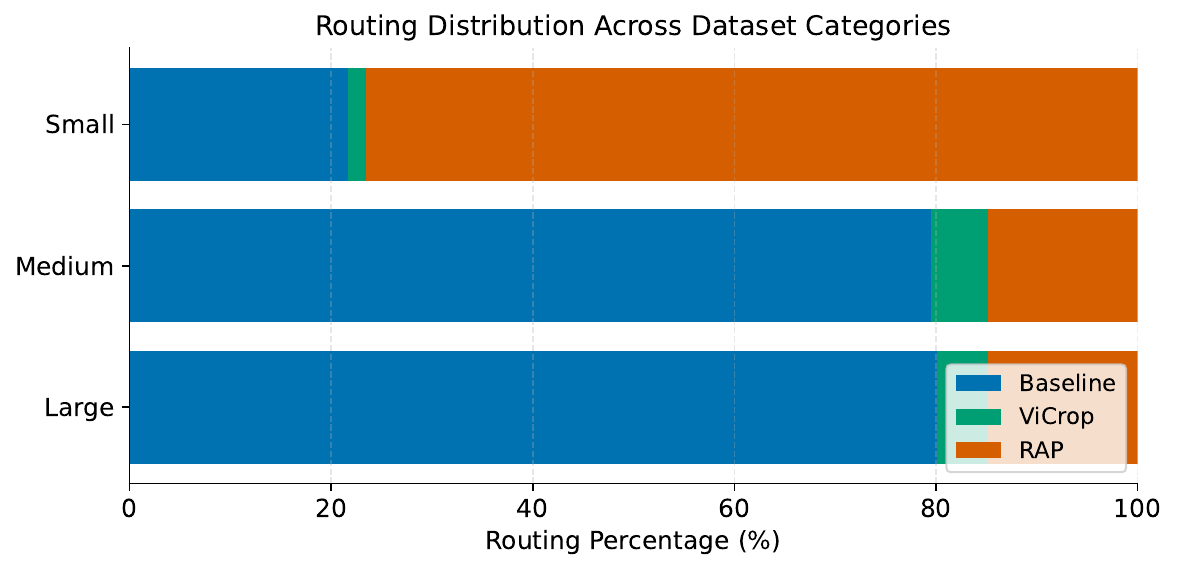}
        \caption{Qwen3-VL-2B routing distribution}
        \label{fig:qwen_routing_distribution}
    \end{subfigure}
    
    \caption{
    Routing distribution across dataset categories for different perception strategies.
    % Blue denotes global perception (Baseline), green denotes attention-based retrieval (ViCrop), and orange denotes rigid patch-based retrieval (RAP).
    }
    
    \label{fig:combined_router_distribution_graphs}
    % \vspace{-5pt}
\end{figure}

\subsection{Routing Decision Analysis}
Figure~\ref{fig:combined_router_distribution_graphs} shows the routing distribution of ViRGo. As we can observe, for fixed-resolution models such as LLaVA-v1.5-7B, the router strongly favors RAP for small objects while increasingly selecting global perception for large objects. A similar trend appears on dynamic-resolution models such as Qwen3-VL-2B, but with a stronger preference toward the global perception as the native models are already  more capable on various object sizes. This suggests that ViRGo allocates expensive crop-based inference only when necessary.

\subsection{Ablation on the Zero-Shot Features}

\begin{table}[h]
\centering
\small
\setlength{\tabcolsep}{6pt}
\renewcommand{\arraystretch}{}

\begin{tabular}{lcccc}
\toprule
\textbf{Feature} &
\textbf{\textit{Small}}&
\textbf{\textit{Medium}}&
\textbf{\textit{Large}}&
\textbf{\textit{W-Avg}$\uparrow$} \\
\midrule
(1) &
52.9	&
63.3	&
74.2	&
62.0	
\\	
(2) &
52.8	&
\textbf{64.1}	&
72.7	&
61.7
\\	
(3) &
50.7	&
63.9	&
75.0	&
61.4	
\\	
(1), (2)  & 
53.5	&
63.3	&
74.2	&
62.2	
\\
(2), (3)& 
53.0	&
62.7	&
74.2	&
61.8	
 
\\
(1), (3) & 
53.5	&
63.3	&
74.2	&
62.2	
\\

\midrule
\textbf{\textit{ViRGo}} & 
\textbf{58.3} & 
63.3 &
\textbf{75.0} &
\textbf{63.2} 
\\
\bottomrule
\end{tabular}

\caption{Comparison of and Accuracy between different feature ablation.}
\label{tab:feature_ablation}
\end{table}
We analyze the contribution of three routing features: (1) model confidence, (2) object size, and (3) image resolution. As shown in Table~\ref{tab:feature_ablation}, using a single feature or pairwise combinations yields limited performance , with weighted averages ranging from 61.4\% to 62.2\%. In contrast, the full ViRGo router achieves the best overall performance at 63.2\%, including a substantial improvement on small objects (+4.8\% over the best ablation). These results show that model confidence, object scale, and image quality provide complementary signals for robust adaptive routing.

\section{Conclusion}
In this work, we identify the resolution-context trade-off in three visual perception paradigms ---global, attention-based, and retrieval-based--- and propose ViRGo, a simple and lightweight yet effective router that decides whether to perform retrieval-based processing based on zero-shot features extracted from the model. Our experiments show ViRGo improves performance and efficiency by avoiding unnecessary computation. ViRGo's effectiveness suggests a promising direction toward more adaptive and resource-efficient visual reasoning systems, where inference strategies are dynamically selected according to the visual demands of the input rather than fixed a priori.

\section{Limitations}

Although ViRGo improves the accuracy--efficiency trade-off through adaptive routing, several limitations remain. The current router primarily relies on coarse geometric and confidence-based signals. While simple and effective for distinguishing small- versus large-object scenarios, these mainly capture geometric aspects of visual perception, leaving other factors that may influence routing decisions---such as relational reasoning, compositional understanding, or multi-object interactions--- unexplored. This limitation is further reflected in the substantial gap between ViRGo and the Oracle routing performance, suggesting that the current routing signals do not yet fully characterize the underlying perception difficulty. Furthermore, ViRGo currently routes among a fixed set of predefined and representative perception strategies (global perception, ViCrop, and RAP). More diverse strategies could further improve the routing's performance, which we leave for future studies.

% For future work, we believe extending ViRGo's binary decision boundary to a continuous one is a promising direction. One could frame this as a regression problem, where the router predicts the optimal number of retrieval patches or visual tokens directly from the heuristic scores. This would require defining a more complex optimization objective beyond binary VQA success and could potentially lead to even more fine-grained adaptive compute. We leave this exciting extension for future exploration.

% \section*{Acknowledgments}

% We thank our advisors, collaborators, and lab members for their valuable feedback and discussions throughout this project. We also acknowledge the open-source community and prior works on visual retrieval augmentation and multimodal reasoning that inspired this research. Computational resources and support from our institution are gratefully appreciated.

% Additional elements were taken from the formatting instructions of the \emph{International Joint Conference on Artificial Intelligence} and the \emph{Conference on Computer Vision and Pattern Recognition}.
%  ABOVE HERE IS MAIN CONTENT (MAX 8 pages)

\newpage

\bibliography{custom}

\clearpage
\appendix

\section*{Appendix Overview}

This appendix provides additional details on the experimental setup, evaluation protocol, and efficiency analysis for ViRGo. 
Appendix~\ref{detailed_implementation} describes the dataset construction and router implementation details, including the train/dev/test split and the lightweight XGBoost configuration used for adaptive routing. 
Appendix~\ref{sec:appendix} defines the object-scale evaluation protocol used to group examples into small-, medium-, and large-object cases across datasets. 
Appendix~\ref{app:time} reports the complete inference-time results across all evaluated backbones, providing further evidence that ViRGo improves the accuracy--efficiency trade-off by avoiding unnecessary retrieval when global perception is sufficient.

\section{ViRGo Detailed Setup}
\label{detailed_implementation}
\paragraph{Dataset Setup.} To evaluate generalization, we combine data from multiple VQA datasets. The data is first split 20\% into training, the remaining data is then split 20\% for validation, and the remaining for test sets, with label balancing applied where possible. The final training set contains approximately 558 samples.

\paragraph{Implementation Details.} We implement the router using a lightweight gradient-boosted decision tree (XGBoost). We use shallow trees (max depth = 2) with regularization. The model is trained using multiclass log loss, with validation data used for early stopping. We find that this simple model is sufficient to achieve good routing performance with minimal computational overhead.

\section{Evaluation Protocol}
\label{sec:appendix}
We evaluate performance across different object scales:

\begin{itemize}
    \item \textbf{Small-object cases:} Adaptive retrieval improves accuracy over base VQA. High-resolution datasets include V*Bench, and HR-Bench4K; low-resolution includes GQA bin-0.
    
    \item \textbf{Medium-object cases:} GQA bin-1, bin-2, bin-3, and TextVQA.
    
    \item \textbf{Large-object cases:} Adaptive retrieval avoids unnecessary zoom and outperforms always-zoom RAP baselines. This includes GQA bin-4 to bin-7.
\end{itemize}

\section{Inference Time Analysis}
\label{app:time}

\begin{table}[h]
\centering
\small
\setlength{\tabcolsep}{6pt}
\renewcommand{\arraystretch}{}

\begin{tabular}{lrrrr}
\toprule
\textbf{Method} &
\textbf{\textit{Small}}&
\textbf{\textit{Medium}}&
\textbf{\textit{Large}}&
\textbf{\textit{Total}$\downarrow$} \\
\midrule

LLaVA-v1.5-7B &
541	&
399	&
395	&
1335	
\\

- ViCrop & 
4074	&
1026	&
934	&
6034	
\\

- RAP & 
71506	&
11252	&
6427	&
89185	
 
\\

\rowcolor{gray!12}
+ \textbf{ViRGo} & 
39374	&
3341	&
1911	&
44626	
\\

\hdashline

LLaVA-v1.5-13B &
642	&
618	&
488	&
1748	
\\

- ViCrop & 
5135	&
2167	&
1470	&
8773	
\\

- RAP & 
77503	&
10079	&
6108	&
93690	
 
\\

\rowcolor{gray!12}
+ \textbf{ViRGo} & 
75247	&
4667	&
2115	&
82039	
\\

\hdashline
LLaVA-ov-0.5B &
1080	&
156	&
104	&
1340	
\\

- ViCrop & 
18438	&
1721	&
1080	&
21239	
\\

- RAP & 
67961	&
7913	&
5278	&
81152	
 
\\

\rowcolor{gray!12}
+ \textbf{ViRGo} & 
50182	&
1517	&
682	&
52381	
\\

\hdashline
Qwen3-VL &
519	&
148	&
105	&
773	
\\

- ViCrop & 
1359	&
458	&
339	&
2156	
\\

- RAP & 
47549	&
2047	&
1057	&
50564	
 
\\

\rowcolor{gray!12}
+ \textbf{ViRGo} & 
42195	&
515	&
349	&
43058	
\\

\bottomrule
\end{tabular}

\caption{Comparison of total inference time between methods (s). ViRGo includes both routing and generation time.}
\label{tab:Total_run_time}
\end{table}

We provide the complete inference-time results in Table~\ref{tab:Total_run_time}. 
All numbers report total wall-clock inference time in seconds over the corresponding evaluation subsets. 
For ViRGo, the reported time includes both the initial global forward pass used to extract routing features and the final response generation after the selected perception strategy is executed.

Global perception is consistently the fastest strategy, since it requires only a single forward pass over the original image. 
ViCrop introduces additional overhead due to attention-based cropping, but remains substantially faster than RAP. 
RAP is the most expensive strategy across all backbones because it performs iterative patch retrieval and evaluates multiple zoomed-in regions. 
In contrast, ViRGo substantially reduces the cost of always-on patch retrieval by invoking expensive crop-based processing only when the router predicts that retrieval is beneficial.

This behavior is most evident on medium- and large-object subsets, where ViRGo frequently routes samples to global perception or ViCrop rather than RAP. 
For example, on LLaVA-v1.5-7B, ViRGo reduces total inference time from 89,185 seconds with RAP to 44,626 seconds, a 50.0\% reduction. 
On LLaVA-ov-0.5B, ViRGo reduces total inference time from 81,152 seconds to 52,381 seconds, a 35.5\% reduction. 
The savings are smaller but still present for LLaVA-v1.5-13B and Qwen3-VL, where ViRGo reduces RAP inference time by 12.4\% and 14.8\%, respectively. 
These results support the main finding that adaptive routing improves the accuracy--efficiency trade-off by avoiding unnecessary high-resolution retrieval, especially when the target object is already sufficiently visible in the global image.

\end{document}